\documentclass[10pt,twocolumn,letterpaper]{article}
\usepackage{wacv}
\usepackage{times}
\usepackage{epsfig}
\usepackage{graphicx}
\usepackage{amsmath}
\usepackage{amssymb}
\usepackage{caption}
\usepackage{subcaption}
\usepackage{MnSymbol}
\makeatletter
\@namedef{ver@everyshi.sty}{}
\makeatother
\usepackage{tikz}
\usetikzlibrary{calc,patterns,angles,quotes}
\usepackage[ruled,vlined]{algorithm2e}
\def\etal{\textit{et al.}}

\usepackage[pagebackref=true,breaklinks=true,colorlinks,bookmarks=false]{hyperref}

\newcommand{\R}{{\mathbb R}}

\SetKwInOut{KwIn}{Input}
\SetKwInOut{KwOut}{Output}
\SetKw{KwRet}{return}
\SetKw{Return}{return}

\newtheorem{theorem}{Theorem}[section]

\newtheorem{definition}[theorem]{Definition}
\newtheorem{definition*}{Definition}

\usepackage{color}

\def \R {\mathbb{R}}

\def\wacvPaperID{1024} 

\wacvfinalcopy 

\ifwacvfinal
\fi

\ifwacvfinal
\usepackage[breaklinks=true,bookmarks=false]{hyperref}
\else
\usepackage[pagebackref=true,breaklinks=true,colorlinks,bookmarks=false]{hyperref}
\fi

\begin{document}

\title{A Riemannian Framework for Analysis of Human Body Surface}


\author{Emery Pierson\\
Univ. Lille, CNRS, Centrale Lille, UMR 9189 CRIStAL, F-59000 Lille, France\\
{\tt\small emery.pierson@univ-lille.fr}
\and
Mohamed Daoudi\\
IMT Nord Europe, Institut Mines-Télécom, Univ. Lille, Centre for Digital Systems, F-59000 Lille, France\\
Univ. Lille, CNRS, Centrale Lille, Institut Mines-Télécom, UMR 9189 CRIStAL, F-59000 Lille, France\\
{\tt\small mohamed.daoudi@imt-nord-europe.fr}
\and
Alice-Barbara Tumpach\\
Univ. Lille, CNRS, UMR 8524 Laboratoire Painlev\'e, F-59000 Lille, France\\
Wolfgang Pauli Institut, Vienna, Austria\\
{\tt\small alice-barbora.tumpach@univ-lille.fr}
}

\maketitle
\ifwacvfinal
\thispagestyle{empty}
\fi
\begin{abstract}
We propose a novel framework for comparing 3D human shapes under the change of shape and pose. This problem is challenging since 3D human shapes vary significantly across subjects and body postures. We solve this problem by using a Riemannian approach. Our core contribution is the mapping of the human body surface  to the space of metrics and normals. We equip this space with a family of Riemannian metrics, called Ebin (or DeWitt) metrics. We treat a human body surface as a point in a "shape space" equipped with a family of Riemannian metrics. The family of metrics is invariant under rigid motions and reparametrizations; hence it induces a metric on the "shape space" of surfaces. Using the alignment of human bodies with a given template, we show that this family of metrics allows us to distinguish the changes in shape and pose. The proposed framework has several advantages. First, we define a family of metrics with desired invariance properties for the comparison of human shape. Second, we present an efficient framework to compute geodesic paths between human shape given the chosen metric. Third, this framework provides some basic tools for statistical shape analysis of human body surfaces. Finally, we demonstrate the utility of the proposed framework in pose and shape retrieval of human body.
\end{abstract}


\maketitle
\section{Introduction}
\label{sec:Introduction}

Human shape analysis is an important area of research with a wide applications in vision, graphics, virtual reality, product design and avatar creation. 
While 3D human shapes are usually represented as 3D surfaces, human bodies
vary significantly across two important properties: shape (or subjects identity) and body postures (or body pose). These variations make human body shape analysis a challenging problem. 
In this paper, we seek a framework for human shape analysis which provides: (i) a shape metric to quantify shape and pose differences (ii)  a full pipeline for generating deformations and shape interpolation; and
(iii) a shape summary, a compact representation of human shapes in terms of the center (mean of human shapes).  


\section{Related Work}

The main tasks in human shape analysis can be divided into representing, comparing, deforming and summarizing human shapes. A common theme in the
literature has been to represent human surfaces by certain geometrical features, such as HKS~\cite{SunOG09}, WKS ~\cite{Aubry2011} and ShapeDNA~\cite{ReuterWP06}. The readers can refer to recent surveys ~\cite{pickup_shape_2016, PRLian13} for an extensive review and comparison of such descriptors. Their structure does not allow for more complex tasks such as interpolation or statistical shape analysis.

Recent deep learning approaches try to tackle this problem.
They use a deep neural networks to build "disentangled" latent spaces~\cite{gdvae_2019, zhou20unsupervised}. However those approaches requires training data, while our approach is using purely geometric information. 
\\
Our approach falls within the class of elastic shapes analysis. In this section we cover methods from this family that are more closely related to ours. Kurtek~\etal~\cite{Kurtekpami2012}) and Tumpach~\etal~(\cite{tumpach_gauge_2016} propose the quotient of the space of embeddings of a fixed surface $S$ into $\R^3$ by the action of the orientation-preserving diffeomorphisms of $S$ and the group of Euclidean transformations, and provide this quotient with the structure of an infinite-dimensional manifold. We can then define and use Riemannian metrics on this manifold to measure the distance between two given shapes as well as to interpolate between them by computing a geodesic that joins them. However, the computational costs of this approach are high.
Another recent approach is that of {\it square root normal fields} or SRNF in which different embeddings and immersions of the surface $S$ modulo translations are described by points in a Hilbert space, and both rotations in $\R^3$ as well as reparametrizations of the surfaces translate into orthogonal transformations in the Hilbert space ~(\cite{JermynEccv2012}). However, the SRNF map is neither injective nor surjective and there exist different shapes having the same SRNF. In addition, as observed by Su el al. \cite{Su_2020_CVPR_Workshops}, the resulting distance can be viewed as an extrinsic distance obtained by embedding the space of parametrized surfaces in a linear space. 

Most of the above approaches use a spherical parameterization of 3D objects, while we propose to use a human template as a parametrization, and take some advantages of the recent developments of static and dynamic human datasets such as SMPL and FAUST.

\subsection{Main Contributions}
In this paper, we present a comprehensive Riemannian framework for analyzing human bodies, in the process of dealing with the change in shape and pose. Unlike some past works, instead of using a general parameterization of human body surfaces, we propose to use a human template and to align the human surfaces to this template. The human body surface is represented by the normal and the induced surface metric. Using the metric on the space of normals and the Ebin metric on the space of Riemannian metrics, a family of metrics is proposed to compare shapes and poses of a human body.  To our best knowledge, this is the first demonstration of the use of this metric in human body shape analysis. We will show also for the first time, that this family of metrics takes into account the intrinsic and extrinsic geometry of human bodies. Additionally, we present an efficient framework  to compute geodesic between given human body surfaces under the chosen metric. We provide some basic tools for statistical shape analysis of human body surfaces. These tools help us to compute an average human body. To evaluate our approach, we conduct extensive experiments on multiple datasets. The experimental results show that the proposed family of Riemannian metrics classifies correctly the shapes and the poses. The experimental results show also that our proposed framework provides better geodesics than the state-of-the-art Riemannian framework.



\section{Mathematical Framework and Background }
\label{sec:MathBackground}
\subsection{Notation}
 Given a reference human being $\mathcal{T}$ (also called a template in the sequel), we will represent a human shape $S$ with an embedding $f~:\mathcal{T}\rightarrow\mathbb{R}^3$ such that the image $f~(\mathcal{T})$ equals $S$  . The map $f$ is an embedding onto a human shape $f(\mathcal{T})$. 
The function $f$ is also called a \textit{correspondence} between the template $\mathcal{T}$ and the human shape $f(\mathcal{T})$.

Recall that a map $f~:\mathcal{T}\rightarrow\mathbb{R}^3$ is an \textit{embedding} when:
(1)  $f$ is smooth, in particular small variations on the template $\mathcal{T}$ correspond to small variations on the human shape $f(\mathcal{T})$ (2)  $f$ is an immersion, i.e. at each point of the human shape  $f(\mathcal{T})$ one can define the normal  (resp. tangent) space to the surface  of the human body
 as subspace of $\mathbb{R}^3$, and (3) $f$ is an homeomorphism onto its image, i.e. points on $f(\mathcal{T})$ that look close
in $\mathbb{R}^3$ are images of close points in $\mathcal{T}$. We define the space of all registered human shapes as
$$\mathcal{H} := \{f~:\mathcal{T}\rightarrow \mathbb{R}^3, f \textrm{~is~an~embedding}\}.$$
It is often called the \textit{pre-shape space} since human bodies with the same shape but 
different correspondences with the template may correspond to different points in $\mathcal{H}$. 
The set $\mathcal{H}$ is a manifold, as an open subset of the linear space $\mathcal{C}^{\infty}(\mathcal{T}, \mathbb{R}^3)$ of smooth functions from 
$\mathcal{T}$ to $\mathbb{R}^3$. 
The tangent space to $\mathcal{H}$ at $f$, denoted by $T_f\mathcal{H}$, is therefore just $\mathcal{C}^{\infty}(\mathcal{T}, \mathbb{R}^3)$.

The shape preserving transformations can be expressed as group actions on $\mathcal{H}$.
The group $\mathbb{R}^3$ with addition as group operation acts on $\mathcal{H}$, 
by \textit{translations}~: $(v, f) \mapsto f+v$, for $v\in \mathbb{R}^3$ and $f\in \mathcal{H}$.
The group $\textrm{SO}(3)$ with matrix multiplication as group operation  acts on $\mathcal{H}$, 
 by \textit{rotations}~: $(O, f)\mapsto O f$, for $O\in \textrm{SO}(3)$ and $f\in \mathcal{H}$. 
Finally, 
the group $\Gamma := \textrm{Diff}^+(\mathcal{T})$ consisting of diffeomorphisms which preserve the orientation of $\mathcal{T}$ acts also on $\mathcal{H}$,
by \textit{reparameterization}~: $(\gamma, f)\mapsto f\circ\gamma^{-1}$, for $\gamma\in \textrm{Diff}^+(\mathcal{T})$ and $f\in\mathcal{H}$.
The use of $\gamma^{-1}$, instead of $\gamma$, ensures that the action is from left and, since the action of $\textrm{SO}(3)$ is also 
from left, one can form a joint action of $G := \textrm{Diff}^+(\mathcal{T})\times  \textrm{SO}(3)$ on $\mathcal{H}$.
In this paper, the translation group is taken care of by using a translation-independent metric.
Therefore, in the following we will 
focus only on the reparameterization group $\Gamma$ and on the rotation group $\textrm{SO}(3)$. 

\subsection{Shape Space of aligned Human bodies}




Given a group $G$ acting on $\mathcal{H}$, the elements in $\mathcal{H}$ obtained by following a fix 
registered human body $f\in\mathcal{H}$ when 
acted on by all elements of $G$ is called the \textit{equivalence class} of $f$ under the action of $G$,
and will be denoted by $[f]$. 
In particular, when $G$ is the reparameterization group $\Gamma :=\textrm{Diff}^+(\mathcal{T})$, the orbit of $f\in \mathcal{H}$ is characterized by the  human shape $f(\mathcal{T}) = S$, 
i.e. the elements in $[f] = \{f\circ \gamma^{-1} \textrm{~for~}\gamma \in \Gamma\}$ are all possible registrations of $S$. 
The quotient space $\mathcal{H}/G$ is called \textit{shape space} and is defined as follows.
\begin{definition}
The \textit{shape space}  $\mathcal{S}$ is the set of (oriented) human bodies in $\mathbb{R}^3$, which are diffeomorphic to $\mathcal{T}$, 
modulo rotation. It is isomorphic to the quotient space of the pre-shape space  
$\mathcal{H}$ by the human motion-preserving group $G := \textrm{Diff}^+(\mathcal{T})\times \textrm{SO}(3)$~:
$\mathcal{S} = \mathcal{H}/G$. 
\end{definition}

In this paper, each human body surface is aligned to a given template $\mathcal{T}$. This means that for any equivalence class $[f]\in \mathcal{H}/G$ a preferred correspondence with the template is chosen. This alignment is anatomically meaningful (for instance the finger tips of the template correspond to the finger tips of the other human bodies (See Figure~3 in the supplementary material). The set of aligned human bodies will be denoted by $\mathcal{S}_0$ and is the space of interest in the present paper. Since the correspondance with the template is chosen in a smooth way, the shape space $\mathcal{S}$ is diffeomorphic to the manifold of aligned human bodies $\mathcal{S}_0$. Mathematically this alignment is called a section $\mathcal{S}_0$  of the fiber bundle $\Pi~:\mathcal{H}\rightarrow \mathcal{H}/G$. (More details are given in supplementary material).

\section{Riemannian Analysis of aligned Human Shapes}
\label{sec:RiemHuman}
Next, we describe our approach to construct the metric between two elements of $\mathcal{S}_0$ and the ``optimal” deformation from one human surface
to another. Since human surfaces are represented as elements of $\mathcal{S}_0$, a natural formulation
of ``optimal” is to consider the two corresponding elements in $\mathcal{S}_0$ and to construct a geodesic connecting them in $\mathcal{S}_0$.

\subsection{Elastic Riemannian Metric}

 Consider a parameterized surface $f~: \mathcal{T} \rightarrow \mathbb{R}^3$. Denote by $g$ the pull-back of the Euclidian metric of 
$\mathbb{R}^3$
and by $n$ the unit normal vector field (Gauss map) on $S = f(\mathcal{T})$. 


%
We consider the following relationship between parameterized
surfaces on one hand and the product space of metrics and normals on the other~:
\begin{equation}\label{Phi}
\begin{array}{lclc}
\Phi~:& \mathcal{S}_0 &\longrightarrow &\operatorname{Met}(\mathcal{T}) \times \mathcal{C}^{\infty}(\mathcal{T}, \mathbb{S}^2)\\
& f & \longmapsto &  (g,  {n}).
\end{array}
\end{equation}

It follows from  the fundamental theorem of surface theory (see Bonnet's Theorem in ~\cite{DoCarmo} for the local result, Theorem 3.8.8 in ~\cite{Klingenberg} or Theorem 2.8-1 in ~\cite{Ciarlet}  
for the global result) that two parameterized surfaces $f_1$ and $f_2$ having the same representation $(g, n)$ differ at most by a translation (and 
rotation for $g$). This theorem implies that we can represent a surface 
by its induced metric $g $ and the unit normal field $n$, for the purpose of analyzing its shape. 
We will not loose any information about the shape of a surface $f$ if we represent
it by the pair $(g,n)$. The induced metric $g$ captures the intrinsic shape, while the normal $n$ captures the extrinsic geometry of shape. The numerical computation of the metric $g$ is included in the supplementary material.

\subsection{The Manifold of Metrics on $\mathcal{T}$ and its Geodesic Distance}
The space of positive-definite Riemannian metrics on $\mathcal{T}$ will be denoted by $\operatorname{Met}(\mathcal{T})$. Once we have selected a Riemannian metric for a human body, it is a point in the infinite-dimensional manifold $\operatorname{Met}(\mathcal{T})$. We will equip the infinite-dimensional space of all Riemannian metrics with a diffeomorphism-invariant Riemannian metric, called the Ebin (or DeWitt) metric~\cite{Ebin1970,PhysRev.160.1113}, as suggested by~\cite{Su_2020_CVPR_Workshops}.
The Riemannian metric on the tangent space is defined by:
\begin{equation}
    ((\delta g, \delta g))_{g} = \int_{\mathcal{T}} \text{Tr}(g^{-1}\delta g_0 g^{-1}\delta g_0) + \lambda \text{Tr}(g^{-1}\delta g)^2 \mu_g
\end{equation}
where $\delta g_0 = \delta g - \frac{1}{2} \text{Tr}(g^{-1}\delta g)g$ is called the traceless part of $\delta g$, and where $\mu_g$ denotes the volume form defined by $g$. 


The following theorem, from~\cite{Su_2020_CVPR_Workshops}, presents the geodesic distance between two metrics $g_1$ and $g_2$ in  (the completion of) $\operatorname{Met}(\mathcal{T})$ for any choice of  $\lambda$.
\begin{theorem} \label{th1}
Let $g_{1}, g_{2} \in \operatorname{Met}(\mathcal{T})$. The square of the geodesic distance for the family of metrics is

$d^{\lambda}\left(g_{1}, g_{2}\right)^{2}=\int_{\mathcal{T}} d ^{\lambda,\mathrm{Sym}}\left(g_{1}(x), g_{2}(x)\right)^{2} d x$

where\\

$\begin{array}{l}
d ^{\lambda,\mathrm{Sym}}\left(g_{1}(x), g_{2}(x)\right)^{2} \\
\quad=16 \lambda\left(s_{1}^{2}(x)-2 s_{1}(x) s_{2}(x) \cos (\theta(x))+s_{2}^{2}(x)\right)
\end{array}$

with\\

$\begin{array}{l}
s_{1}(x)=\sqrt[4]{\operatorname{det}\left(g_{1}(x)\right)}, \quad s_{2}(x)=\sqrt[4]{\operatorname{det}\left(g_{2}(x)\right)} \\
\theta(x)=\min \left\{\pi, \frac{\sqrt{\lambda^{-1} \operatorname{tr}\left(K_{0}^{2}(x)\right)}}{4}\right\} \\
K(x)=\left\{\begin{array}{l}
0 \text { if either } g_{1}(x) \text { or } g_{2}(x) \text { is degenerate } \\
g_{1}(x) \log \left(g_{1}(x)^{-1} g_{2}(x)\right) \text { else }
\end{array}\right. \\
K_{0}(x)=K(x)-\operatorname{tr}\left(g_{1}^{-1}(x) K(x)\right) g_{1}(x)
\end{array}
$
\end{theorem}

\begin{theorem} \label{th2} 
Let $a, \lambda, c$, three positive real numbers.
We equip the space $\operatorname{Met}(\mathcal{T}) \times C^{\infty}\left(\mathcal{T}, \mathbb{S}^{2}\right)$ with the following Riemannian metric:
\begin{equation}
\label{metric_g_n}
\begin{array}{l}
    (((\delta g, \delta n), (\delta g, \delta n)))_{g, n} =\\
    \quad\quad
    a\left(\int_{\mathcal{T}} \text{Tr}(g^{-1}\delta g_0 g^{-1}\delta g_0) + \lambda \text{Tr}(g^{-1}\delta g)^2 \mu_g\right) \\
    \quad\quad + c \int_{\mathcal{T}} \langle \delta n, \delta n \rangle dx 
    \end{array}
\end{equation}
Let $f_1, f_2 \in \mathcal{S}_0$ and $\Phi\left(f_{1}\right)=\left(g_{1}, n_{1}\right), \Phi\left(f_{2}\right)=\left(g_{2}, n_{2}\right)$. Define a distance function $d_{\mathcal{S}_0}$ on $\mathcal{S}_0$ by 
\begin{equation}
d_{\mathcal{S}_0}^{a, \lambda, c}\left(f_{1}, f_{2}\right) := d\left(\Phi\left(f_{1}\right), \Phi\left(f_{2}\right)\right)
\end{equation}
where $d$ is the geodesic distance in the space $\operatorname{Met}(\mathcal{T}) \times C^{\infty}\left(\mathcal{T}, \mathbb{S}^{2}\right)$. Then the square of the distance $d_{\mathcal{S}_0}$ between $f_1$ and $f_2$, with parameters $a, \lambda, c$, is given by 

\begin{equation}
\label{equ:distance}
d_{\operatorname{\mathcal{H}}}^{a, \lambda, c}\left(f_{1}, f_{2}\right)^{2}= a d^{\lambda}\left(g_{1}, g_{2}\right)^{2}+c\int_{\mathcal{T}} d_{\mathbb{S}^2}(n_{1}(x), n_{2}(x))^{2} dx
\end{equation}
where $d^\lambda$ is given by Theorem~\ref{th1} and $d_{\mathbb{S}^2}(n_{1}(x), n_{2}(x)) = \arccos{\left\langle n_1(x), n_{2}(x)\right\rangle}$ is the geodesic distance on $\mathbb{S}^2$.
\end{theorem}

\subsection{Computation of Geodesics}
As mentioned above, an important advantage of our Riemannian approach over many past papers is its ability to compute not only the distance between two human surfaces but also  the geodesics or the deformations between shapes. The computation of geodesics requires the minimization of an energy. In ~\cite{tumpach_gauge_2016} the path-straightening method is used to find critical
points of the energy functional. Starting with an arbitrary path, the method consists of iteratively deforming (or “straightening”) the path in the opposite direction of the gradient, until the path converges to a geodesic. The problem would then be a problem of optimization on the set of vertices of the shape. However, this can lead to numerical instabilities. We
will use another, more stable approach ~\cite{su_shape_2019}. In this approach, after choosing a time step $\frac{1}{T}, T\in \mathbb{N}$, the path is set to the linear path (initialization) on which we add a sum of deformations:

\begin{equation}
\begin{aligned}
&f\left(t_{0}\right)=f_{0}, \quad f\left(t_{T}\right)=f_{1} \\
&f\left(t_{i}\right)=\left(1-t_{i}\right) f_{0}+t_{i} f_{1}+\sum_{j} \alpha_{ij} \mathcal{D}_j
\end{aligned}
\label{eq:Geodesic}
\end{equation}

Where $\mathcal{D}_j$ is an orthogonal basis of $N_{\mathcal{D}}$ plausible deformations gathered beforehand.
The computation of the geodesic requires the minimization of the energy functional  $E(\alpha)$, defined by:
\begin{equation}
E(\alpha) = \int_0^{1}  \left(\left(\frac{d\Phi(f(t))}{dt},\frac{d \Phi(f(t))}{dt}\right)\right)_{\Phi(f(t))}dt
\label{eq:Energy}
\end{equation}
with $\alpha \in \mathbb{R}^{(T-2)*N_{\mathcal{D}}}$ the vector containing all $\alpha_{ij}$ presented in equation~\ref{eq:Geodesic}, and $((.,.))_{\Phi(f(t))}$ being the pullback by $\Phi$ of the Riemannian metric \ref{metric_g_n} on  $\operatorname{Met}(\mathcal{T}) \times C^{\infty}\left(\mathcal{T}, \mathcal{S}^{2}\right)$. 


To find the optimal coefficients $\alpha$, similar to ~\cite{su_shape_2019}, we employ the  Broyden–Fletcher–Goldfarb–Shanno (BFGS) method~\cite{fletcher_practical_1987}, implemented in the SciPy  library~\cite{2020SciPy-NMeth} where we calculate the gradient using the automatic differentiation feature of PyTorch library~\cite{pytorch2019_9015}.

\paragraph{Basis Deformations}\label{deformation_basis}
In ~\cite{Kurtekpami2012}, \cite{LagaPAMI2017}, ~\cite{Su_2020_CVPR_Workshops}, ~\cite{tumpach_gauge_2016}, spherically parameterization of 3D objects is used and spherical harmonics are computed to define the set of deformations. However, human surfaces
will require a large number of basis elements to
achieve high accuracy and capture all the human surface
details. In addition, in the case of human shapes, we are using a human template as a parametrization and there are several publicly available dynamic human shapes that can be used to build a PCA basis of deformations.

In our case to build such real deformations, we use the publicly Dynamic FAUST dataset~\cite{Bogo:CVPR:2014}, which contains motions registered to the template $\mathcal{T}$. 
10 individuals (5 males, 5 females) perform 14 different motions, sampled at the rate of 60 frame per second.
Given a set of motions, we collect deformations by gathering differences from the sequences.
Let $(m_1, ..., m_T) \in \mathcal{S}_0$ be a motion available in the dataset. We define the small deformations that we collect from the motions as the family $(m_{n\tau + \tau} - m_{n\tau})_n$, with $\tau$ being a time interval chosen manually, fixed to 10 frames ($\simeq$160 ms).
Thus, given a set of training samples, we can compute its PCA basis. In our experiments, the number of PCA basis elements required is of the order of 100.

Note that, by construction, adding a deformations of the basis of deformation to a aligned human shape will not destroy the alignment with the template.


\begin{algorithm}
\SetAlgoLined
\KwIn{the source and target surfaces $f_1$ and $f_2$, $a, \lambda, c$ the parameter of the elastic metric}
\KwOut{$f_{\text{geo}}$: the geodesic connecting $f_1$ and $f_2$}
 1: Initialize $\alpha_{ij}=0$ and $f(t_i)$ by linear path\;
 2: Define the energy functional $E(\alpha)$ in an automatic differentiation framework (PyTorch here), that computes the gradient value $\nabla_\alpha E$ along the functional value\;
 3: Minimize $E$ with respect to $\alpha$ with a BFGS implementation (SciPy \textit{BFGS} or \textit{L-BFGS-B}), that uses the gradient $\nabla_\alpha E$\;
 4: Set the geodesic to be: $f_{\text{geo}}(t_i) = t_i f_0 + (1-t_i) f_1 + \sum_j \alpha_{ij} \mathcal{D}_j $\;
 5: \Return{the final geodesic $f_{\text{geo}}$}
 \caption{Computation of Geodesics}
 \label{algo:Geodesics}
\end{algorithm}

\section{Statistical Analysis of Human Shapes}

We are interested in defining a notion of “mean” for a given set of human shape. 
Let $f_1,\dots  f_n$ be a set of human shapes. The mean of a set of human shapes is the human shape that is as close as possible to all of the human shapes in the set of human shapes, under the distance metric defined by Equation~\ref{equ:distance}. This is known as the Karcher mean and is defined as the human shape that minimizes the sum of squared distances to all of the human shape in the given human shape. In order to find the Karcher mean one can define the following functional:
\begin{equation}
    \mathcal{V}: \mathcal{S}_0 \rightarrow \mathbb{R}, \mathcal{V}(f)= \sum_{i=1}^{n} d\left(f_{i}, f \right)^{2}
\end{equation}
That is differentiable with the distance previously computed. We initialize the Karcher mean as $f_1$ and set it to be the sum of $f_1$ with a linear combination of deformations:
$$
\bar{f} = f_1 + \sum_j \beta_j \mathcal{D}_i
$$
The functional to minimize becomes:
$$
\mathcal{W}(\beta) = \mathcal{V}(f_1 + \sum_i \beta_i \mathcal{D}_i)
$$


\begin{algorithm}
\SetAlgoLined

\KwIn{$f_1,\dots  f_n$ a set of human body, $a, \lambda, c$ the parameter of the elastic metric}
\KwOut{$\bar{f}$: Karcher mean }
 1: Initialize $\alpha=0$ and $\bar{f}=f_1$ by the first shape in the set\;
 2: Define the Karcher mean functional $\mathcal{W}^{a, \lambda, c}(\beta)$ in an automatic differentiation framework (PyTorch here) that computes the gradient value $\nabla_\beta \mathcal{W}$ along the functional value\;
 3: Minimize $\mathcal{W}$ with respect to $\beta$ with a BFGS implementation (SciPy \textit{BFGS} or \textit{L-BFGS-B}), that uses the gradient $\nabla_\beta \mathcal{W}$\;
 4: Set the Karcher mean to be: $\bar{f} = f_1 + \sum_i \beta_i \mathcal{D}_i$\;
 5: \Return {Karcher mean}
 \caption{Karcher Mean of Human Shapes}
 \label{algo:KarcherMean}
\end{algorithm}


\section{Experiments}
\subsection{Assessment of the Family of Elastic Metrics}

\begin{figure}\label{figure_pose}
    \centering
    \includegraphics[width=1\linewidth]{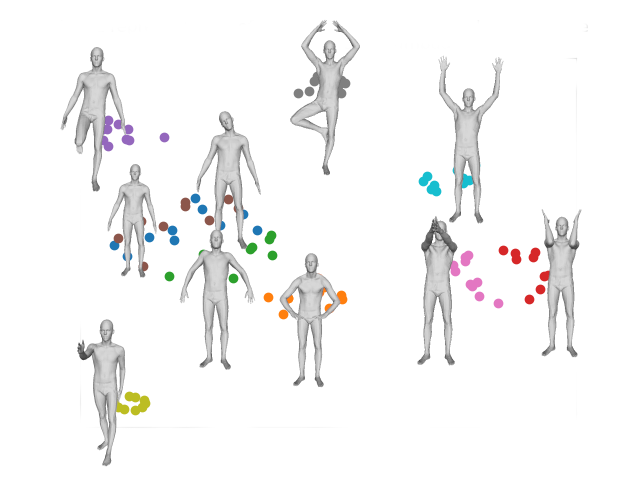}
    \caption{2D visualization of the FAUST dataset by our method using t-SNE algorithm based on the metric from equation~\ref{equ:distance}. The metric parameters are set to $a=1, \lambda=0.0001, c=0$. Each color represents a class of pose and a class representative is also displayed.}
    \label{fig:TSNE_pose}
\end{figure}

\begin{figure}\label{figure_shape}
    \centering
    \includegraphics[width=1\linewidth]{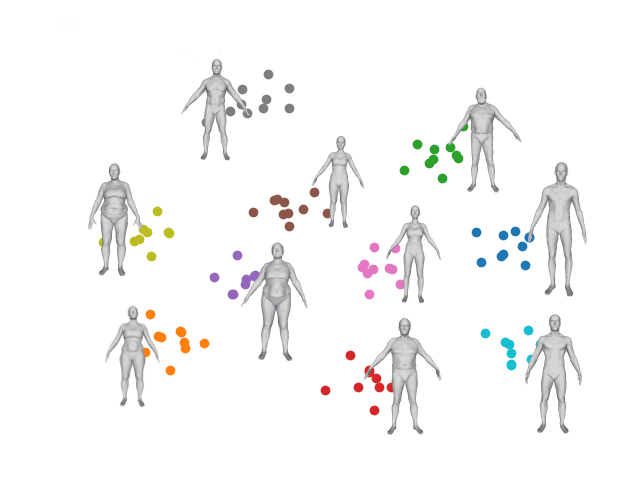}
    \caption{2D visualization of the FAUST dataset by our method using t-SNE algorithm based on the metric from equation~\ref{equ:distance}. The metric parameters are set to $a=0, \lambda=0, c=1$. Each color represents a class of shape and a class representative is also displayed.}
    \label{fig:TSNE_shape}
\end{figure}

To further assess the pertinence of the family of elastic distances defined in Equation~\ref{equ:distance} in human shape and pose analysis, we measured pairwise distances of the metric on the registrations present in the FAUST dataset~\cite{Bogo:CVPR:2014}. It contains 10 individuals (5 males, 5 females) in 10 different poses. We present in Figure~\ref{fig:TSNE_pose} and ~\ref{fig:TSNE_shape} 2D visualizations of the dataset using the t-Distributed Stochastic Neighbor Embedding (t-SNE) algorithm~\cite{vandermaaten08a}. 

The Figure~\ref{fig:TSNE_pose} clearly evidences that the 3D human with similar poses belong to very close distributions. These results show the assumption that given $a=0, \lambda=0, c=1$ (normal field $L_2$ metric), the metric is preserved under shape change, and could be used in pose and motion analysis application~\cite{luo_spatio-temporal_2016, Veinidis19}.
The figure ~\ref{fig:TSNE_shape} shows that 3D human with similar shape belong to very close distribution. These results states the assumption that given $a=1, \lambda=0.0001, c=0$, the metric is preserved under pose change, and could be used in many shape analysis application  approaches ~\cite{pickup_shape_2016} and ~\cite{PRLian13}.


\subsection{Geodesics and Karcher Mean}

We performed a number of experiments using human surfaces of same and different persons under a variety of pose and shape, and studied the resulting geodesic paths. 

\begin{figure*}
    \centering
    \begin{subfigure}{\linewidth}
    \centering
    \includegraphics[width=0.69\linewidth]{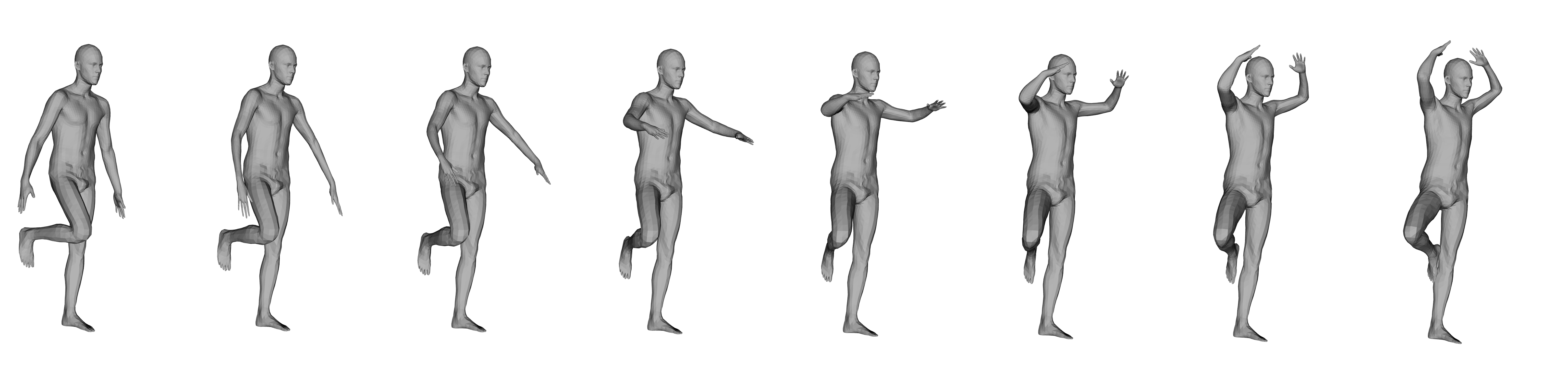}
    \includegraphics[width=0.28\linewidth]{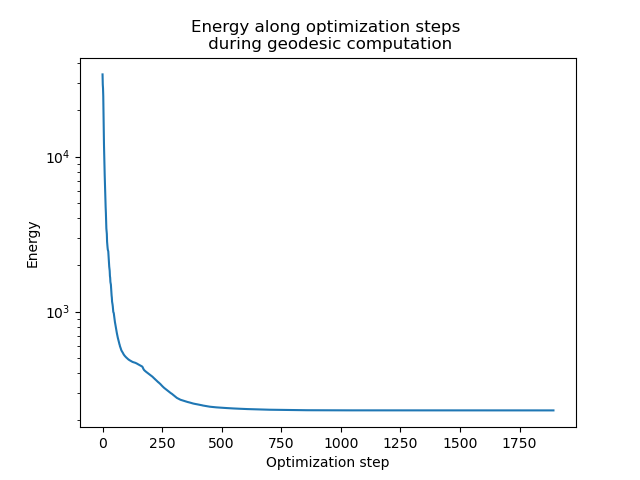}
    \subcaption{}
    \end{subfigure}
    ~
    \begin{subfigure}{\linewidth}
    \centering
    \includegraphics[width=0.69\linewidth]{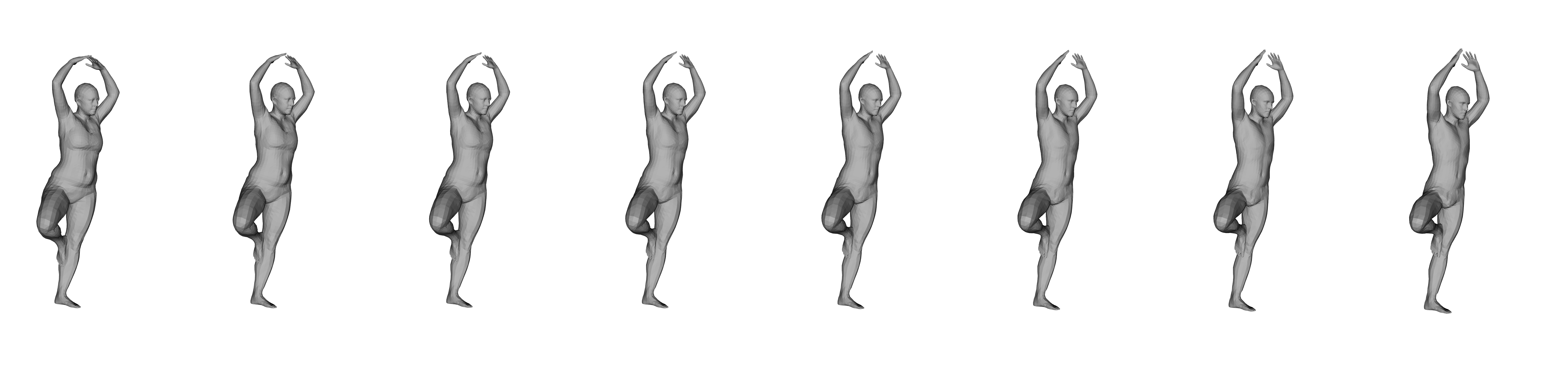}
    \includegraphics[width=0.28\linewidth]{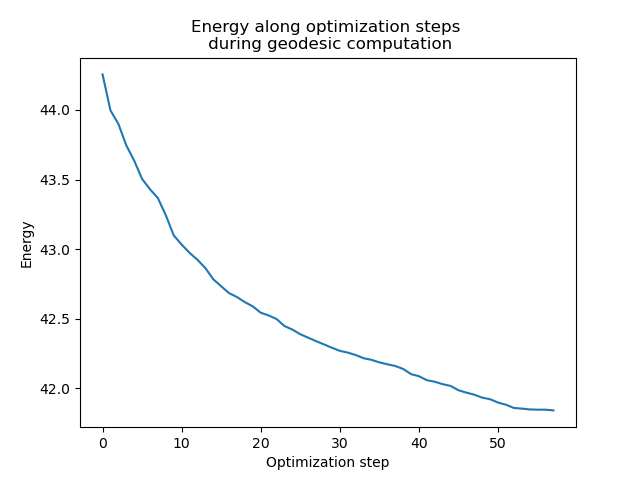}
    \subcaption{}
    \end{subfigure}
     ~
    \caption{Examples of geodesic path between $f_1$ and far left and $f_2$ far right: (a)  with metric parameters (a=1, $\lambda=1$, c=0), (b) with metric parameters (a=0, $\lambda=0$, c=1). The corresponding energy evolution during optimization are displayed on the right. Computation time was respectively 3min31s and 10.6s.}
    \label{fig:all_geos}
\end{figure*}

Figure~\ref{fig:all_geos} shows the geodesic
path between $f_1$ (shown in far left) and $f_2$ (shown in far right). Drawn in between are human surfaces denoting equally spaced points along the geodesic path. In terms of the Riemannian
metric chosen, these paths denote the optimal deformations in going from the first human body to the second and the path lengths quantify the amount of deformations.
For this experiment, we also provide a curve of the energy, available right to the paths, which shows that the energy decreases smoothly with time. 

For the first path, the change in the pose induces small changes in shape. We thus want to minimize the shape change along the path, which would set the extrinsic parameters $c=0$. We find that $a=1, \lambda=1$ gives the best visual results.

The second path is a path with change in shape.  We thus want to minimize the pose change along the path, which would set parameters $a=\lambda=0$, and the normal parameter $c=1$.
 
The geodesic computation were made on a computer setup with Intel(R) Xeon(R) Bronze 3204 CPU @ 1.90GHz, and a Nvidia Quadro RTX 4000 8GB GPU. The computation time of the different geodesics took less than 5 mins.

\begin{figure}
    \centering
    \includegraphics[width=0.7\linewidth]{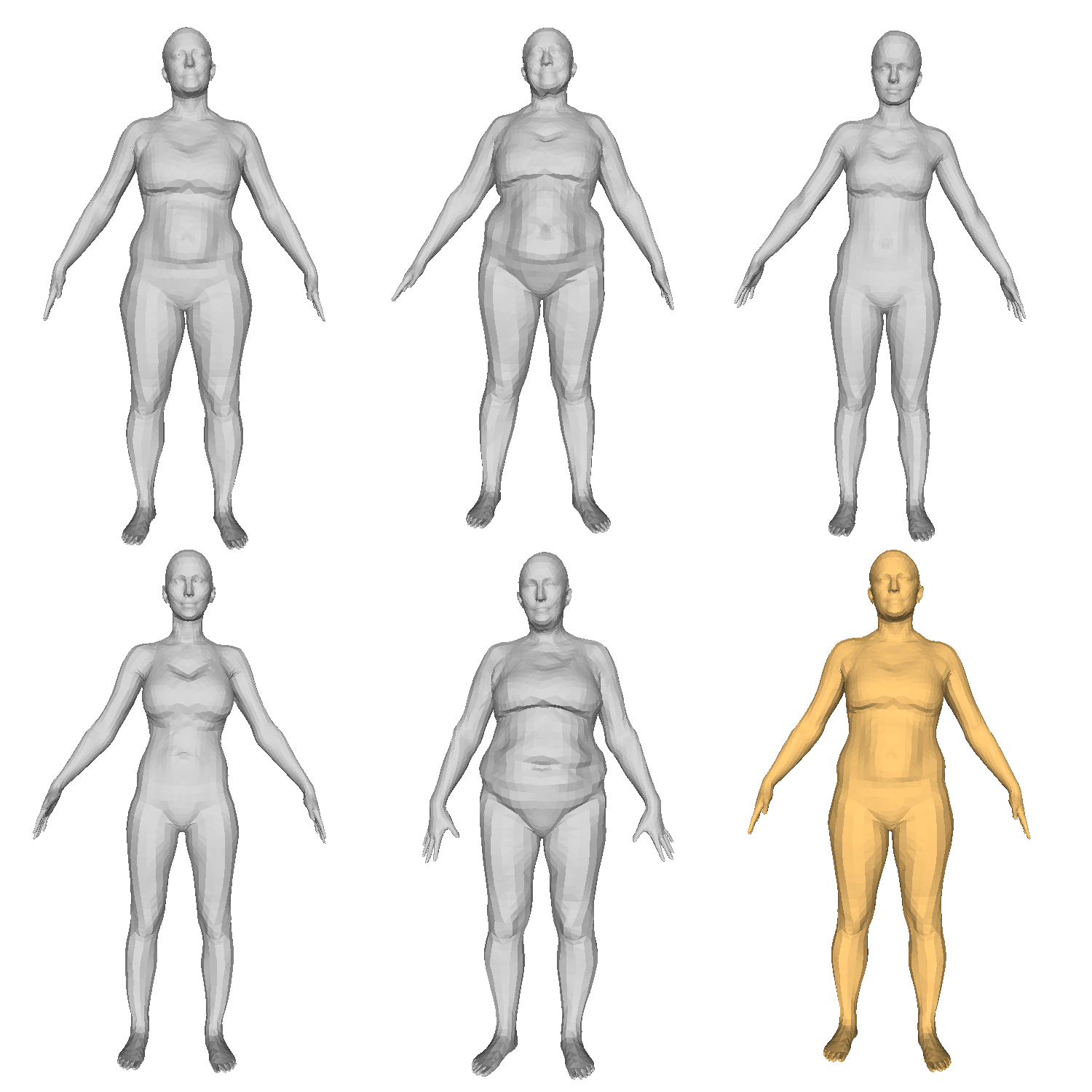}
    \caption{Karcher mean (yellow) for a five different people relatively to the distance with metric parameters $(a=0, \lambda=0, c=1)$.}
    \label{fig:karcher_shape}
\end{figure}

\begin{figure*}
    \centering
    \begin{subfigure}{\linewidth}
        \centering
        \includegraphics[width=0.9\linewidth]{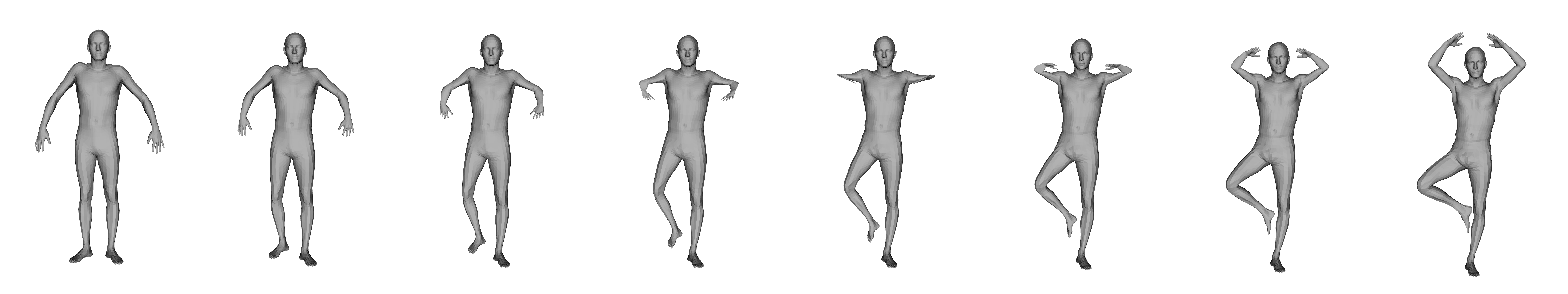}
        \subcaption{Linear geodesic path}
        \label{fig:geo_lin}
    \end{subfigure}
    ~
    \centering
    \begin{subfigure}{\linewidth}
        \centering
        \includegraphics[width=0.9\linewidth]{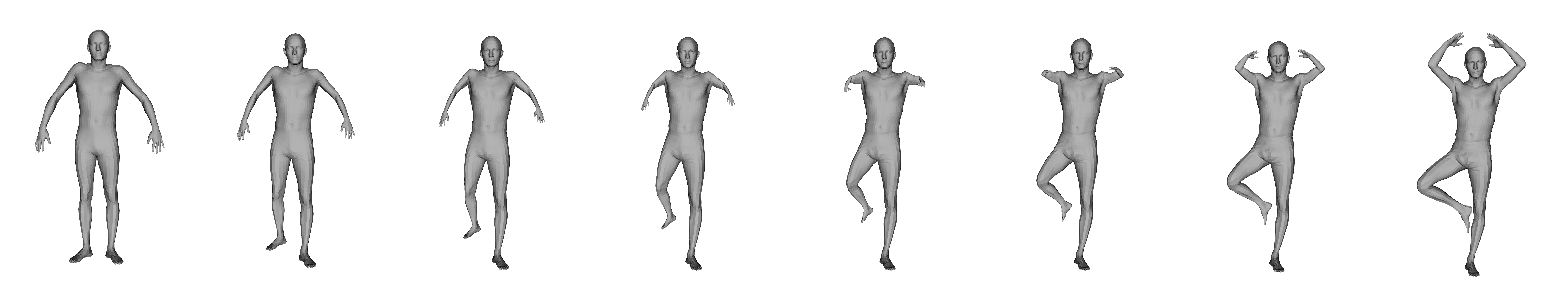}
        \subcaption{Geodesic computed with SRNF}
        \label{fig:geo_srnf}
    \end{subfigure}
    ~
    \centering
    \begin{subfigure}{\linewidth}
        \centering
        \includegraphics[width=0.9\linewidth]{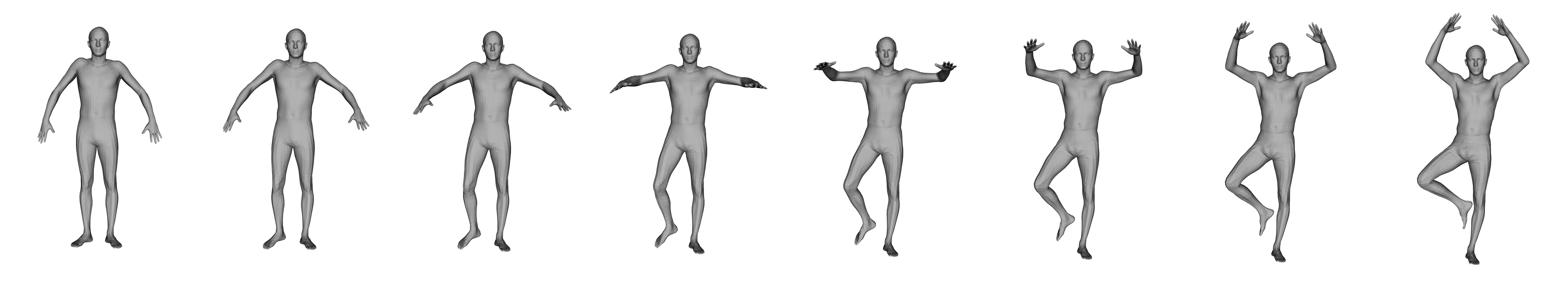}
        \subcaption{Geodesic computed in SMPL space}
        \label{fig:geo_smpl}
    \end{subfigure}
    ~
    \centering
    \begin{subfigure}{\linewidth}
        \centering
        \includegraphics[width=0.9\linewidth]{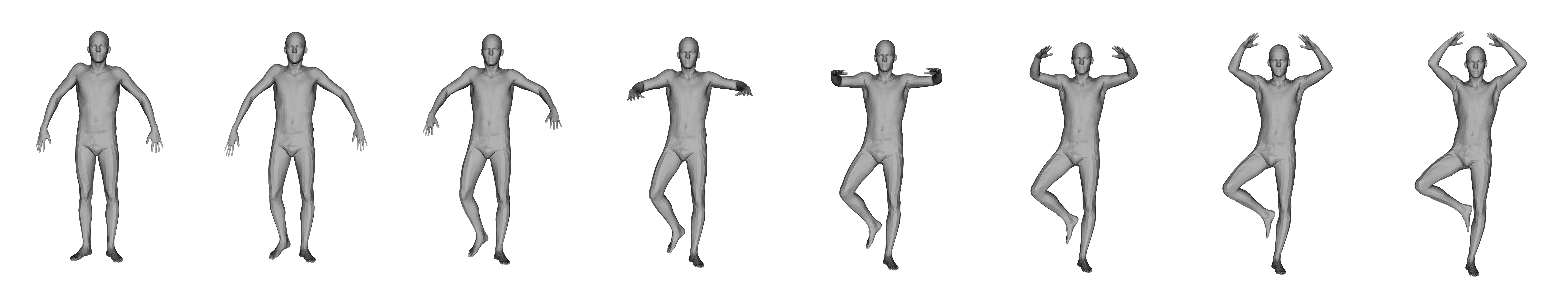}
        \subcaption{Geodesic computed with our approach, metric parameters are set to $a=1, \lambda=1, c=0$. Computation time was 3min10s.}
        \label{fig:geo_us}
    \end{subfigure}
    ~
    \caption{Comparison of our approach with different
frameworks. We observe that the linear initial path is challenging, while the SRNF path induces distortion in the shape. Finally although the SMPL geodesic is able to keep the shape, we argue that the path of our approach is the most natural path compared to the one proposed by SMPL: the natural deformations between the source and target shape would indeed bend more the elbow. 
}
    \label{fig:geos}
\end{figure*}

An example of Karcher mean is shown in Figure~\ref{fig:karcher_shape}, where five human bodies in the same pose but with different shapes are averaged. 


We also compare the results obtained with
our method to the results using linear geodesic path, SRNF and SMPL descriptors.

\begin{enumerate}
\item The linear geodesic path defined by:
\begin{equation}
\begin{aligned}
&f\left(t_{0}\right)=f_{0}, \quad f\left(t_{T}\right)=f_{1} \\
&f\left(t_{i}\right)=\left(1-t_{i}\right) f_{0}+t_{i} f_{1} 
\end{aligned}
\end{equation}

\item The SRNF  geodesic path is also visualized. This representation has been used to analyze human shapes with interesting results~\cite{LagaPAMI2017, su_shape_2019}. The SRNF is a pointwise representation based on $q = \sqrt{A}n$, where $A = \lVert f_u \times f_v \rVert$ is the area, and $n$ the normal field. We compute the geodesic for the $SRNF$ representation with the same method as presented in this section.

As shown in Figures \ref{fig:geos}(a) and (b), the linear interpolation and SRNF lead to unnatural deformations for human paths. The deformation between surfaces
contains many artifacts and degeneracies.
\item SMPL body model ~\cite{SMPL:2015} : The SMPL model is a human blend shape model. The human shape is presented as a function of $\beta, \theta$, with $\theta$ being the parameters of human body pose, as a cartesian product of axis angle rotation of skeletal joints (21 joints), in axis-angle representation, which lives in $\R^{21*3} = \R^{63}$. $\beta$ are the parameters of the human body shape being the coefficients of linear combination of Principal Component Analysis (PCA) shape decomposition (10 components). After fitting SMPL model to the FAUST dataset, we can compute the corresponding geodesic, using the resulting shapes of the linear path in the SMPL parameter space, see Figure~\ref{fig:geos}. While the deformation propose by the SMPL body model is in some way plausible, we first argue that the pose deformation proposed by SMPL does not bend enough the elbow: this is due to the linear interpolation of the elbow joint angle. In addition, one can observe that the target and sources shapes are slightly different than for our shapes: this is due to the fitting step of SMPL: the resulting shape is the closest shape with plausible SMPL parameters, not exactly the input shape.
\end{enumerate}

In both examples, our approach provides better results.




\section{Application to Pose and Shape Retrieval}
\label{sec:Retrieval}
Here, we demonstrate how the proposed metric can be exploited for 3D human retrieval. Given a 3D human, we look for the similar 3D human in a database.
\subsection{Evaluation Metrics and Comparisons}

We test the usefulness of the family of metrics (Equation~\ref{equ:distance}) in 3D human shape and pose retrieval. We use Nearest neighbor (NN), First-tier (FT), Second-tier (ST) as evaluations criteria.


\noindent \textbf{Comparisons:}
We propose four methods for comparison with our method. The first method GDVAE~\cite{gdvae_2019} is a point cloud variational autoencoder which is trained to disantengle the intrinsic and extrinsic informations of a given shape in the latent space, and propose a latent vector that decomposes in an intrinsic and extrinsic part. We used the FAUST meshes as input of their available trained network and gathered their extrinsic latent vectors, which lives in $\mathbb{R}^{12}$, along with their intrinsic latent vectors, which were for human pose retrieval and shape retrieval respectively. 
The network has been trained on the SURREAL dataset~\cite{varol17_surreal} .\\
The second method  proposed by Zhou et al. ~\cite{zhou20unsupervised} is a mesh autoencoder based on Neural3DMM~\cite{neural3dmm_2019} graph neural network structure. They disantengle the shape and pose in the latent space. 
We apply the FAUST meshes on their available network, trained on AMASS dataset, and use the pose latent vector, which lives in $\mathbb{R}^{112}$ as a descriptor for comparison. 
For human shape, the Area Projection Transform~\cite{giachetti_radial_2012} which won the human shape retrieval challenge~\cite{pickup_shape_2016} is presented. It has been designed for a different goal here, since it is parameterization invariant.
We also compare to the SRNF distance that showed reliable results for pose retrieval.
Finally we use both shape and pose representation from the SMPL body model for the respective retrieval tasks.
\subsection{Experimental results}
In this section, we perform evaluations of our method in FAUST dataset. We evaluate on pose and shape retrieval. 
The evaluation results in Table ~\ref{tab:NN_shape} demonstrate that our method outperforms the
previous state of the art shape retrieval methods in term of NN criteria. The Table ~\ref{tab:NN_pose} shows that the proposed approach provides the best results on pose retrieval in term of FT and ST criteria. We also find that for shape retrieval, the best parameters are $a=1, \lambda<<a$. The computation times for each pairwise distance were $\simeq$70 ms and $\simeq$80 ms for pose and shape retrieval respectively. 
\begin{table}[]
    \centering
    \begin{tabular}{l|c|c|c}
    Repr. & NN & FT & ST \\ \hline \hline 
        GDVAE intrinsic~\cite{gdvae_2019} & 27 & 24.8 & 46.2 \\ \hline
        Zhou et al. shape\cite{zhou20unsupervised} & 42 & 24.8 & 42.8 \\ \hline
        SMPL shape vector & 98 & 72.4 & 86.7 \\ \hline
        APT~\cite{giachetti_radial_2012} & 96 & 86.5 & 96.2 \\ \hline
        Metric (1, 0.0001, 0) & \textbf{100} & \textbf{94.8} & \textbf{97.1}\\ \hline
        
    \end{tabular}
    \caption{FAUST dataset results for shape retrieval}
    \label{tab:NN_shape}
\end{table}

\begin{table}[]
    \centering
    \begin{tabular}{l|c|c|c}
    Repr. & NN & FT & ST \\ \hline \hline 
        GDVAE extrinsic~\cite{gdvae_2019} & 60 & 38.0 & 54.2 \\ \hline
        Zhou et al. pose\cite{zhou20unsupervised} & 82 & 69.2 & 83.4\\ \hline
        SMPL pose vector & 80 & 84.4 & 95.2 \\ \hline
        SRNF & 73 & 77.7 & 94.4 \\ \hline 
        Metric (0, 0, 1) & \textbf{85} & \textbf{88.3} & \textbf{97.6} \\ \hline
    \end{tabular}
    \caption{FAUST dataset results for pose retrieval}
    \label{tab:NN_pose}
\end{table}
\section{Conclusion}
\label{sec:Conclusion}
In this paper we have proposed a novel Riemannian framework which allows not only to compute a metric between human bodies under pose and shape changes, but also provides a geodesic path between  human bodies, and  statistical tools (eg. mean of human shape). We have demonstrated the utility of the proposed framework in pose and shape retrieval of human body. The main limitation of our method lies in the requirement of a template.
\section*{Acknowledgments}
This work was supported by the ANR project Human4D ANR-19-CE23-0020. This work was also partially supported by the French State, managed by National Agency for Research (ANR) under the Investments for the future program with reference ANR-16-IDEX-0004 ULNE. The authors thank E.~Klassen, M.~Bauer and Z. ~Su from Department of Mathematics, Florida State University,  for the  discussion on the implementation of geodesic computation. We are grateful to B.~Levy from  Inria Nancy Grand-Est research center for the discussion about the computation of induced metric $g$ on a triangulated mesh.

\newpage
\section*{Appendices}
\appendix
\section{SMPL Template}

\begin{figure}[!h]
    \centering
    \includegraphics[width=0.85\linewidth]{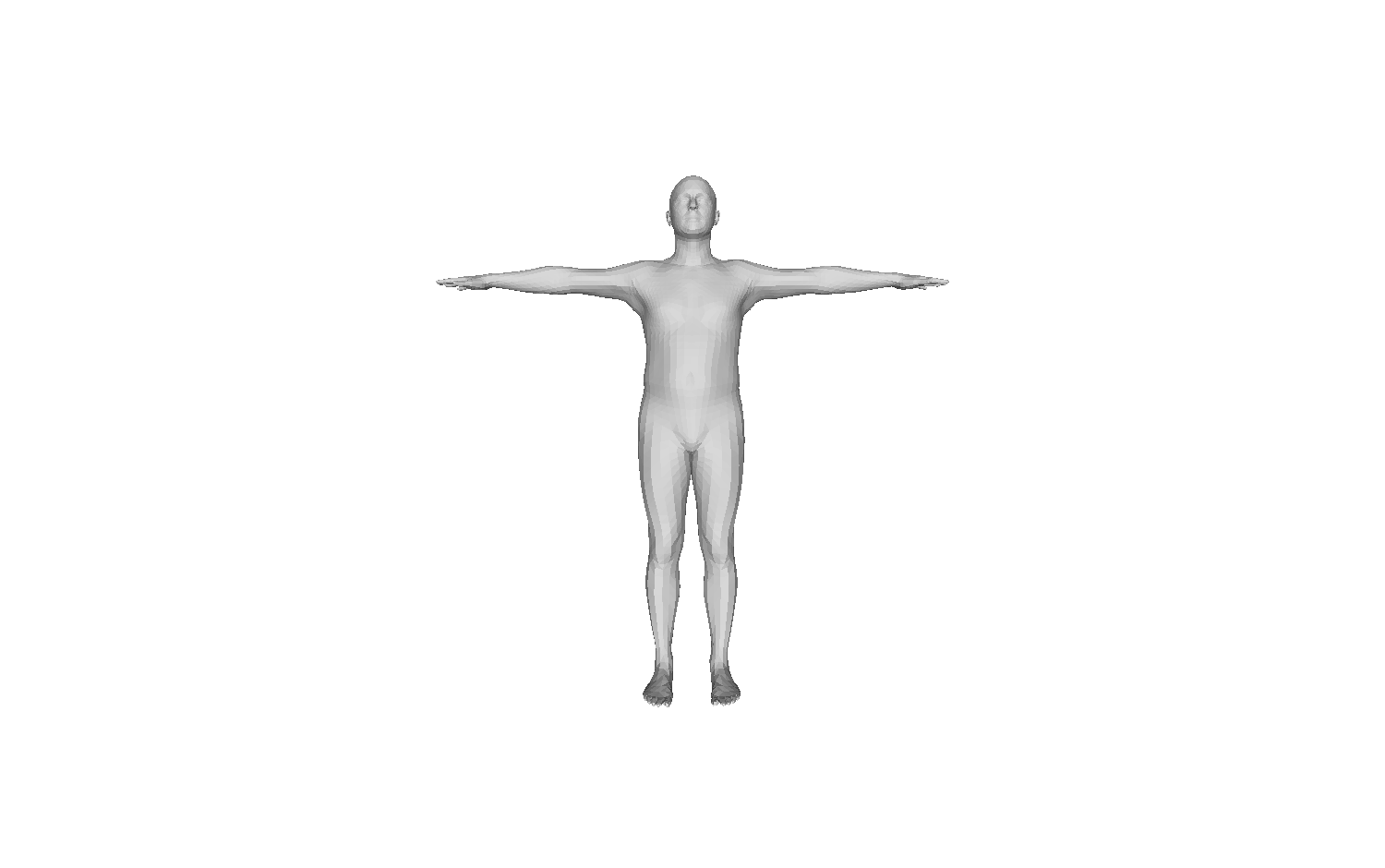}
    \caption{SMPL template}
    \label{fig:smpl_template}
\end{figure}

The Figure \ref{fig:smpl_template} shows the used SMPL ~\cite{SMPL:2015} template $\mathcal{T}$.

\section{Examples from FAUST dataset}
\begin{figure}[!h]
    \centering
    \includegraphics[width=1.1\linewidth]{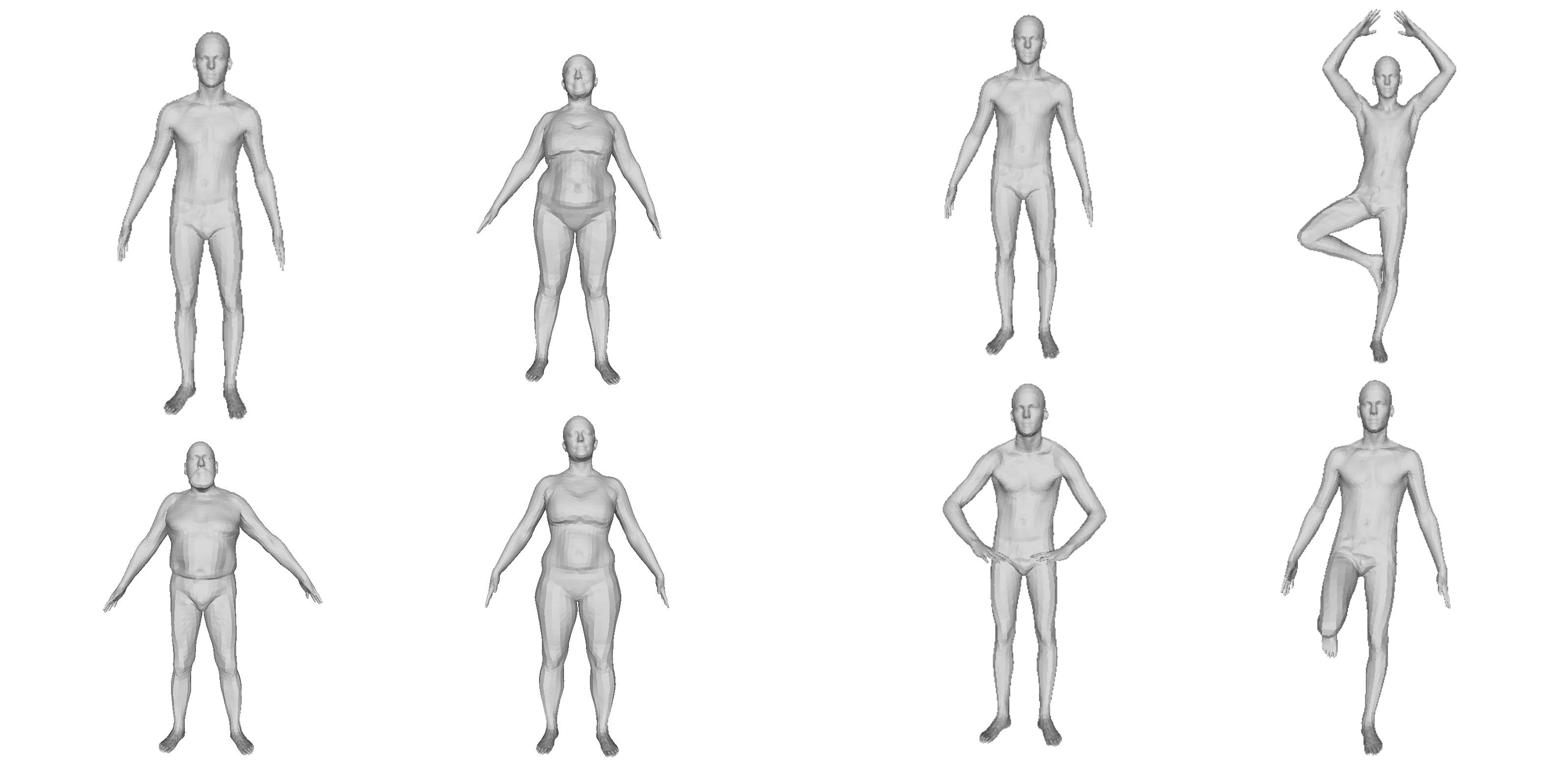}
    \caption{Human shapes from the FAUST dataset.}
    \label{fig:faust_poses}
\end{figure}

The Figure \ref{fig:faust_poses} shows some examples of human shapes from FAUST dataset ~\cite{Bogo:CVPR:2014}. One can see that there is a significant variability in shape and pose.

\section{Shape Space as Section of a Fiber Bundle}

\begin{figure}[ht]
 		\centering
 		\includegraphics[width=0.9\linewidth]{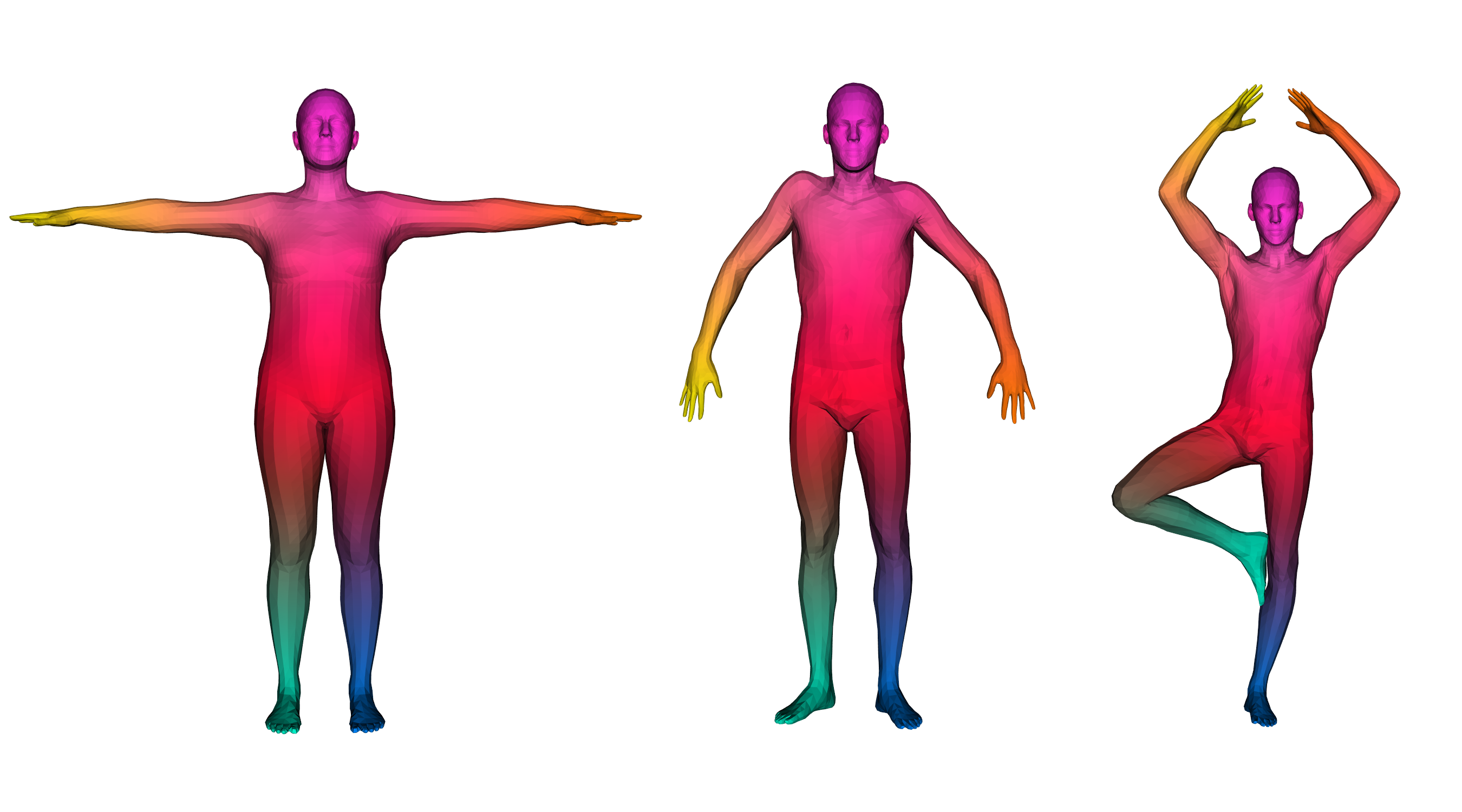}
		\caption{Alignment with the template: The 3 meshes in different poses are displayed with different color on extremities. This validate the choice to work on this particular section of the fiber bundle $\Pi~:\mathcal{H}\rightarrow \mathcal{H}/G$}
		\label{Correspondance}
\end{figure}

In this paper, each human body surface is aligned to a given human template (SMPL template). As illustrated in Figure~\ref{Correspondance},
the geometric features of the template are aligned with geometric feature of the human surface (for instance, the finger tips of the template correspond to the finger tips of the other humans bodies).

\begin{figure}[ht]
 	\centering
 	\begin{tikzpicture}
        \coordinate[label=right:$\mathcal{H}$] (H) at (0,3);
        \coordinate[label=below right:{$G_{f_0}=\Pi^{-1}([f_0])$}] (G) at (2.7, 4);
        \draw[ultra thick]
        (0,0) node [black,label=above right:$\mathcal{S}_0$] {} to [out=5,in=165](3,0);
        \draw[ultra thick]
        (1,2) to [out=5,in=165](4,2);
        \draw[ultra thick]
        (0,0) to [out=85,in=235](1,2);
        \draw[ultra thick]
        (3,0) to [out=85,in=235](4,2);
        
        \draw[thick]
        (0,-5) node [black,label=above right:$\mathcal{H}/G$] {} to (3,-5);
        \draw[thick]
        (1,-3) to (4,-3);
        \draw[thick]
        (0,-5) to (1,-3);
        \draw[thick]
        (3,-5) to (4,-3);

        \coordinate [label=right:$f_0$] (x) at (2.5,1);
        \coordinate (x_g1) at (2.3,1.1);
        \coordinate (x_g2) at (2.1,1.2);
        \coordinate (x_g3) at (1.9,1.3);
        \coordinate [label=below:{$[f_0]$}] (pix) at (2.8, -4);
        \coordinate [label=left:$\Pi$] (pi) at (2.25, -2.7);
        \coordinate [label=right:$s_0$] (pi) at (3, -2.0);
        \fill[black] (x) circle (0.07);
        \fill[gray] (x_g1) circle (0.07);
        \fill[gray] (x_g2) circle (0.07);
        \fill[gray] (x_g3) circle (0.07);
        \fill[black] (pix) circle (0.07);
        
        \draw[black, thick] (2.7, -2) .. controls (2.45,1) .. (2.7, 4);

        \draw[gray, thick] (2.5, -2) .. controls (2.25,1) .. (2.5, 4);
        \draw[gray, thick] (2.3, -2) .. controls (2.05,1) .. (2.3, 4);
        \draw[gray, thick] (2.1, -2) .. controls (1.85,1) .. (2.1, 4);
        
        \draw[->, black, thick, dashed] (2.4, 0.5) .. controls(2.2, -1) .. (2.7, -3.8);
        \draw[->, black, thick] (2.9, -3.8) .. controls(3, -1.5) .. (2.6, 0.5);
        \end{tikzpicture}
 	\caption{Section of the fiber bundle $\Pi~:\mathcal{H}\rightarrow \mathcal{H}/G$: the one-to-one correspondance with the template mesh allows us to work on the corresponding section $\mathcal{S}_0$ as a shape space. The correspondance initially gives the section for $\textrm{Diff}^+({\mathcal{T}})$, but with Procrustes analysis, the section for $\textrm{SO}(3)$ comes straightforwardly.}
	\label{Section_fiber_bundle}
\end{figure}
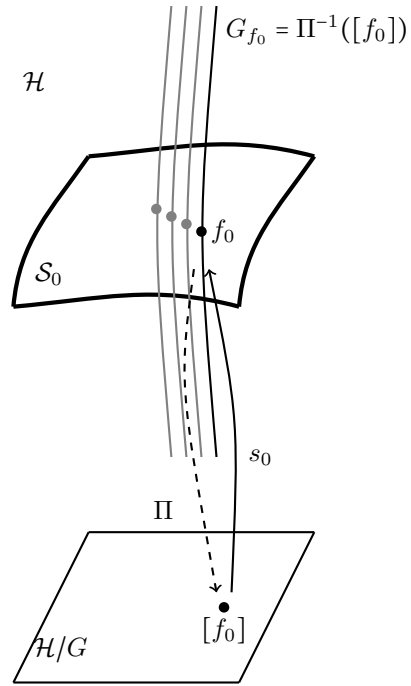

Mathematically the choice of a preferred alignment with the template is called a section $\mathcal{S}_0$  of the fiber bundle $\Pi~:\mathcal{H}\rightarrow \mathcal{H}/G$. A section of $\Pi$ is a (smooth) map assigning to each equivalence class $[f_0]\in  \mathcal{H}/G$ a representative $f_0\in \mathcal{H}$ in this class, i.e. such that $\Pi(f_0) = [f_0]$.  This notion is illustrated in Figure~\ref{Section_fiber_bundle}. The section we are using, i.e. the correspondence with the template, is smooth, thanks to the geometric alignment as explained above.  An illustration of the section is displayed in Figure~\ref{Section_fiber_bundle}.

In this paper, we pull-back the Riemannian metrics that are defined on shape space (see in section 4) on the preferred section $\mathcal{S}_0$ given by the correspondence with the template.

\section{Numerical Computation of the First Fundamental Form $g$}
The computation of the first fundamental $g$  is given by:

$$g=\left(\begin{array}{ll}
\left\langle\frac{\partial f}{\partial u}, \frac{\partial f}{\partial u}\right\rangle & \left\langle\frac{\partial f}{\partial u}, \frac{\partial f}{\partial v}\right\rangle \\
\left\langle\frac{\partial f}{\partial u}, \frac{\partial f}{\partial v}\right\rangle & \left\langle\frac{\partial f}{\partial v}, \frac{\partial f}{\partial v}\right\rangle
\end{array}\right)
$$

It is always computed relatively to a given parameterization. Here the parameterization is given relatively to the template.

The first step is to describe a canonical frame for each template triangle $t_i^T$:
Let $p_1, p_2, p_3$ be the 3 corners of the triangle.
We set the origin of the frame to $0$. We design the local coordinates by $(u, v)$ in the plane denoted by the triangle, that we define with : the coordinates of the triangle corners being $(0, 0), (u_2, 0), (u_3, v_3)$.

Now we need to find the $f$ that maps the triangle $t_i$ in the template to the triangle $t^m_i$ in the destination mesh. Let $q_1, q_2, q_3$ be the 3 corners of the triangle.

In a given $(u, v)$ of $t_i$, the parameterization is of the form:
$$
f(u, v) = \lambda_1 q_1 + \lambda_2 q_2 + \lambda_3 q_3
$$
In order to compute the derivatives, we need to derive $f$. The derivative is given by:
$$
\frac{\partial f(u, v)}{\partial u} = \frac{\partial \lambda_1}{\partial u}  q_1 + \frac{\partial \lambda_2}{\partial u} q_2 + \frac{\partial \lambda_3}{\partial u} q_3
$$
$$
\frac{\partial f(u, v)}{\partial v} = \frac{\partial \lambda_1}{\partial v}  q_1 + \frac{\partial \lambda_2}{\partial v} q_2 + \frac{\partial \lambda_3}{\partial v} q_3
$$
Since the triangle are in correspondence, only the $\lambda_i$ -- the barycentric coordinates -- depend on $(u,v)$.
Given $(u, v)$, those $\lambda_i$ are defined as the solution of the following equation (see~\cite{levy_hdr} for similar calculations):
$$
\begin{pmatrix}
0 & u_2 & u_3 \\
0 & 0 & v_3 \\
1 & 1 & 1 \\
\end{pmatrix} \begin{pmatrix}
\lambda_1 \\ \lambda_2 \\ \lambda_3
\end{pmatrix} = \begin{pmatrix}
u \\ v \\ 1
\end{pmatrix}
$$
The solution is straightforward and given by:
$$
\frac{\partial f(u, v)}{\partial u} =  \frac{1}{u_2}(q_2 - q_1)
$$
$$
\frac{\partial f(u, v)}{\partial v} =  \frac{u_3}{v_3 u_2}(q_1 - q_2) + \frac{1}{v_3}(q_3 - q_1)
$$
The values of $u_2$ is simply the length of the first edge of the triangle ($l_1$). 
$u_3$ and $v_3$ are the projection of first and second edge on the $u$-axis and $v$-axis. So $v_3$ is the height H of the triangle (relative to first edge as the basis), and $\frac{u_3}{v_3}$ is $\tan(\theta)$, where $\theta$ is the angle between first edge and second edge:
$$
\frac{\partial f(u, v)}{\partial u} =  \frac{1}{l_1}(q_2 - q_1)
$$
$$
\frac{\partial f(u, v)}{\partial v} =  \frac{\tan{\theta}}{l_1}(q_1 - q_2) + \frac{1}{H}(q_3 - q_1)
$$
\begin{figure}
\centering
    \begin{tikzpicture}
    \coordinate[label=left:$p_1$] (origo) at (0,0);
    \coordinate (pivot) at (1,5);

    \fill[black] (origo) circle (0.05);
    \draw[thick,gray,->] (origo) -- ++ (1.5, 0) node[black, below]{$l_1$} -- ++(2.5,0) node[black,right] {$u$};
    \draw[thick,gray,->] (origo) -- ++(0,4) node (mary) [black,left] {$v$};

    \draw[thick,fill={rgb:red,0.5;green,1;blue,1.25}] (origo) -- (2, 3) node (p2) [black,label=above:$p_3$] {} -- (3, 0) node (p3) [black,label=below:$p_2$] {} -- (origo);
    \fill[black] (p2) circle (0.05);
    
    \fill[black] (p3) circle (0.05);
    \node[gray] (H) (2,0) {};
    \draw[thick,dashed,gray] (p2) -- (2, 1) node [gray, left] {H} -- (2, 0) node[gray] {};
    \pic [draw, ->, "$\theta$", angle eccentricity=1.5] {angle = p3--origo--p2};
  \end{tikzpicture}
  \caption{Ordered triangle with the corresponding local reference frame. The quantities $l_1, H, \theta$ are useful in the computation of the derivatives.}
\end{figure}
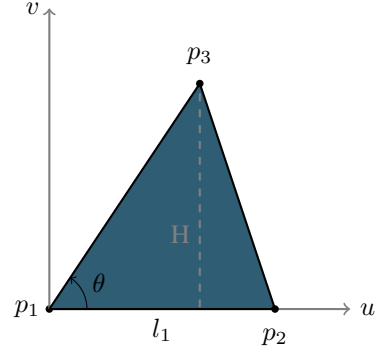




\begin{thebibliography}{10}\itemsep=-1pt

\bibitem{Aubry2011}
Mathieu Aubry, Ulrich Schlickewei, and Daniel Cremers.
\newblock The wave kernel signature: A quantum mechanical approach to shape
  analysis.
\newblock In {\em 2011 IEEE International Conference on Computer Vision
  Workshops (ICCV Workshops)}, pages 1626--1633, 2011.

\bibitem{gdvae_2019}
Tristan Aumentado-Armstrong, Stavros Tsogkas, Allan Jepson, and Sven Dickinson.
\newblock Geometric disentanglement for generative latent shape models.
\newblock In {\em Proceedings of the IEEE/CVF International Conference on
  Computer Vision (ICCV)}, October 2019.

\bibitem{Bogo:CVPR:2014}
Federica Bogo, Javier Romero, Matthew Loper, and Michael~J. Black.
\newblock {FAUST}: Dataset and evaluation for {3D} mesh registration.
\newblock In {\em Proceedings IEEE Conf. on Computer Vision and Pattern
  Recognition (CVPR)}, Piscataway, NJ, USA, June 2014. IEEE.

\bibitem{neural3dmm_2019}
Giorgos Bouritsas, Sergiy Bokhnyak, Stylianos Ploumpis, Michael Bronstein, and
  Stefanos Zafeiriou.
\newblock Neural {3D} {Morphable} {Models}: {Spiral} {Convolutional} {Networks}
  for {3D} {Shape} {Representation} {Learning} and {Generation}.
\newblock {\em arXiv:1905.02876 [cs]}, Aug. 2019.
\newblock arXiv: 1905.02876.

\bibitem{Ciarlet}
P.~G. Ciarlet.
\newblock {\em An Introduction to Differential Geometry with Applications to
  Elasticity}, volume 78-79.
\newblock Kluwer Academic Publishers, 2005.

\bibitem{PhysRev.160.1113}
Bryce~S. DeWitt.
\newblock Quantum theory of gravity. i. the canonical theory.
\newblock {\em Phys. Rev.}, 160:1113--1148, Aug 1967.

\bibitem{DoCarmo}
M.~P. do Carmo.
\newblock {\em An Introduction to Differential Geometry with Applications to
  Elasticity}.
\newblock Prentice-Hall, Inc., Englewood Cliffs, New Jersey, 1976.

\bibitem{Ebin1970}
D.~G. Ebin.
\newblock The manifold of {R}iemannian metrics, in : Global analysis, berkeley,
  calif., 1968.
\newblock {\em Proc. Sympos. Pure Math.}, 15:11--40, 1970.

\bibitem{fletcher_practical_1987}
R.~(Roger) Fletcher.
\newblock {\em Practical methods of optimization}.
\newblock Chichester ; New York : Wiley, 1987.

\bibitem{giachetti_radial_2012}
A. Giachetti and C. Lovato.
\newblock Radial {Symmetry} {Detection} and {Shape} {Characterization} with the
  {Multiscale} {Area} {Projection} {Transform}.
\newblock {\em Computer Graphics Forum}, 31(5):1669--1678, 2012.

\bibitem{JermynEccv2012}
Ian~H. Jermyn, Sebastian Kurtek, Eric Klassen, and Anuj Srivastava.
\newblock Elastic shape matching of parameterized surfaces using square root
  normal fields.
\newblock In {\em ECCV (5)}, pages 804--817, 2012.

\bibitem{Klingenberg}
W. Klingenberg.
\newblock {\em Eine Vorlesung in Differentialgeometrie}.
\newblock Springer Verlag, Berlin, 1973.

\bibitem{Kurtekpami2012}
Sebastian Kurtek, Eric Klassen, John~C. Gore, Zhaohua Ding, and Anuj
  Srivastava.
\newblock Elastic geodesic paths in shape space of parameterized surfaces.
\newblock {\em IEEE Trans. Pattern Anal. Mach. Intell.}, 34(9):1717--1730,
  2012.

\bibitem{LagaPAMI2017}
Hamid Laga, Qian Xie, Ian~H. Jermyn, and Anuj Srivastava.
\newblock Numerical inversion of {SRNF} maps for elastic shape analysis of
  genus-zero surfaces.
\newblock {\em {IEEE} Trans. Pattern Anal. Mach. Intell.}, 39(12):2451--2464,
  2017.

\bibitem{levy_hdr}
Bruno L{\'e}vy.
\newblock {\em {G{\'e}om{\'e}trie Num{\'e}rique}}.
\newblock Habilitation {\`a} diriger des recherches, {Institut National
  Polytechnique de Lorraine - INPL}, Feb. 2008.

\bibitem{PRLian13}
Zhouhui Lian, Afzal Godil, Benjamin Bustos, Mohamed Daoudi, Jeroen Hermans,
  Shun Kawamura, Yukinori Kurita, Guillaume Lavou{\'{e}}, Hien~Van Nguyen,
  Ryutarou Ohbuchi, Yuki Ohkita, Yuya Ohishi, Fatih Porikli, Martin Reuter,
  Ivan Sipiran, Dirk Smeets, Paul Suetens, Hedi Tabia, and Dirk Vandermeulen.
\newblock A comparison of methods for non-rigid 3d shape retrieval.
\newblock {\em Pattern Recognit.}, 46(1):449--461, 2013.

\bibitem{SMPL:2015}
Matthew Loper, Naureen Mahmood, Javier Romero, Gerard Pons-Moll, and Michael~J.
  Black.
\newblock {SMPL}: A skinned multi-person linear model.
\newblock {\em ACM Trans. Graphics (Proc. SIGGRAPH Asia)}, 34(6):248:1--248:16,
  Oct. 2015.

\bibitem{luo_spatio-temporal_2016}
Guoliang Luo, Frederic Cordier, and Hyewon Seo.
\newblock Spatio-temporal segmentation for the similarity measurement of
  deforming meshes.
\newblock {\em The Visual Computer}, 32(2):243--256, Feb. 2016.

\bibitem{pytorch2019_9015}
Adam Paszke, Sam Gross, Francisco Massa, Adam Lerer, James Bradbury, Gregory
  Chanan, Trevor Killeen, Zeming Lin, Natalia Gimelshein, Luca Antiga, Alban
  Desmaison, Andreas Kopf, Edward Yang, Zachary DeVito, Martin Raison, Alykhan
  Tejani, Sasank Chilamkurthy, Benoit Steiner, Lu Fang, Junjie Bai, and Soumith
  Chintala.
\newblock Pytorch: An imperative style, high-performance deep learning library.
\newblock In H. Wallach, H. Larochelle, A. Beygelzimer, F. d\textquotesingle
  Alch\'{e}-Buc, E. Fox, and R. Garnett, editors, {\em Advances in Neural
  Information Processing Systems 32}, pages 8024--8035. Curran Associates,
  Inc., 2019.

\bibitem{pickup_shape_2016}
D. Pickup, X. Sun, P.~L. Rosin, R.~R. Martin, Z. Cheng, Z. Lian, M. Aono,
  A.~Ben Hamza, A. Bronstein, M. Bronstein, S. Bu, U. Castellani, S. Cheng, V.
  Garro, A. Giachetti, A. Godil, L. Isaia, J. Han, H. Johan, L. Lai, B. Li, C.
  Li, H. Li, R. Litman, X. Liu, Z. Liu, Y. Lu, L. Sun, G. Tam, A. Tatsuma, and
  J. Ye.
\newblock Shape {Retrieval} of {Non}-rigid {3D} {Human} {Models}.
\newblock {\em International Journal of Computer Vision}, 120(2):169--193,
  2016.

\bibitem{ReuterWP06}
Martin Reuter, Franz{-}Erich Wolter, and Niklas Peinecke.
\newblock Laplace-beltrami spectra as 'shape-dna' of surfaces and solids.
\newblock {\em Comput. Aided Des.}, 38(4):342--366, 2006.

\bibitem{Su_2020_CVPR_Workshops}
Zhe Su, Martin Bauer, Eric Klassen, and Kyle Gallivan.
\newblock Simplifying transformations for a family of elastic metrics on the
  space of surfaces.
\newblock In {\em Proceedings of the IEEE/CVF Conference on Computer Vision and
  Pattern Recognition (CVPR) Workshops}, June 2020.

\bibitem{su_shape_2019}
Zhe Su, Martin Bauer, Stephen~C. Preston, Hamid Laga, and Eric Klassen.
\newblock Shape {Analysis} of {Surfaces} {Using} {General} {Elastic} {Metrics}.
\newblock {\em arXiv:1910.02045 [math]}, Oct. 2019.
\newblock arXiv: 1910.02045.

\bibitem{SunOG09}
Jian Sun, Maks Ovsjanikov, and Leonidas~J. Guibas.
\newblock A concise and provably informative multi-scale signature based on
  heat diffusion.
\newblock {\em Comput. Graph. Forum}, 28(5):1383--1392, 2009.

\bibitem{tumpach_gauge_2016}
Alice~Barbara Tumpach, Hassen Drira, Mohamed Daoudi, and Anuj Srivastava.
\newblock Gauge {Invariant} {Framework} for {Shape} {Analysis} of {Surfaces}.
\newblock {\em IEEE Transactions on Pattern Analysis and Machine Intelligence},
  38(1):46--59, Jan. 2016.
\newblock arXiv: 1506.03065.

\bibitem{vandermaaten08a}
Laurens van~der Maaten and Geoffrey Hinton.
\newblock Visualizing data using t-{SNE}.
\newblock {\em JMLR}, 9(86):2579--2605, 2008.

\bibitem{varol17_surreal}
G{\"u}l Varol, Javier Romero, Xavier Martin, Naureen Mahmood, Michael~J. Black,
  Ivan Laptev, and Cordelia Schmid.
\newblock Learning from synthetic humans.
\newblock In {\em CVPR}, 2017.

\bibitem{Veinidis19}
Christos Veinidis, Antonios Danelakis, Ioannis Pratikakis, and Theoharis
  Theoharis.
\newblock Effective descriptors for human action retrieval from 3d mesh
  sequences.
\newblock {\em Int. J. Image Graph.}, 19(3):1950018:1--1950018:34, 2019.

\bibitem{2020SciPy-NMeth}
Pauli Virtanen, Ralf Gommers, Travis~E. Oliphant, Matt Haberland, Tyler Reddy,
  David Cournapeau, Evgeni Burovski, Pearu Peterson, Warren Weckesser, Jonathan
  Bright, St{\'e}fan~J. {van der Walt}, Matthew Brett, Joshua Wilson, K.~Jarrod
  Millman, Nikolay Mayorov, Andrew R.~J. Nelson, Eric Jones, Robert Kern, Eric
  Larson, C~J Carey, {\.I}lhan Polat, Yu Feng, Eric~W. Moore, Jake
  {VanderPlas}, Denis Laxalde, Josef Perktold, Robert Cimrman, Ian Henriksen,
  E.~A. Quintero, Charles~R. Harris, Anne~M. Archibald, Ant{\^o}nio~H. Ribeiro,
  Fabian Pedregosa, Paul {van Mulbregt}, and {SciPy 1.0 Contributors}.
\newblock {{SciPy} 1.0: Fundamental Algorithms for Scientific Computing in
  Python}.
\newblock {\em Nature Methods}, 17:261--272, 2020.

\bibitem{zhou20unsupervised}
Keyang Zhou, Bharat~Lal Bhatnagar, and Gerard Pons-Moll.
\newblock Unsupervised shape and pose disentanglement for 3d meshes.
\newblock In {\em European Conference on Computer Vision (ECCV)}, August 2020.

\end{thebibliography}

\end{document}